# Machine Learning-Based Classification Algorithms for the Prediction of Coronary Heart Diseases

Kelvin Kwakye & Emmanuel Dadzie

*Abstract*— Coronary heart disease, which is a form of cardiovascular disease (CVD), is the leading cause of death worldwide. The odds of survival are good if it is found or diagnosed early. The current report discusses a comparative approach to the classification of coronary heart disease datasets using machine learning (ML) algorithms. The current study created and tested several machine-learning-based classification models. The dataset was subjected to Smote to handle unbalanced classes and feature selection technique in order to assess the impact on two distinct performance metrics. The results show that logistic regression produced the highest performance score on the original dataset compared to the other algorithms employed.

In conclusion, this study suggests that LR on a well-processed and standardized dataset can predict coronary heart disease with greater accuracy than the other algorithms.

## I. INTRODUCTION

According to the World Health Organization, 17.9 million people died in 2019 from cardiovascular diseases (CVDs), accounting for 32% of all global deaths [1]. Cardiovascular diseases (CVDs) are a group of disorders of the heart and blood vessels which include coronary heart disease, cerebrovascular disease, peripheral arterial disease, congenital heart disease, etc. Despite advances in healthcare and medicine research, the rate of coronary heart disease (CHD) has been steadily increasing over time, and researchers all over the world have been working to identify the factors that are associated with future risk of coronary heart disease (CHD). Medical doctors' current techniques for predicting and diagnosing heart disease are primarily based on an examination of the medical history, symptoms, and physical reports of the patient. Most often, medical experts struggle to accurately predict a patient's heart disease, whereas they can predict with up to 67% accuracy in some cases [2].

As a result, the medical field requires an automated intelligent system to accurately predict heart disease to help in the decision-making process. This can be accomplished by combining the enormous patient data available in the medical sector with machine learning or deep learning algorithms and intelligent decision-making systems [3]. These big data found in healthcare database repositories, if properly processed, will help to reduce the prevalence of some of these diseases. This will help medical practitioners predict diseases and recommend diagnoses more easily and quickly.

## II. RELATED WORK

Recent years have seen extensive research into machine learning and deep learning algorithms in a variety of fields, including medicine [4-6], transportation [7-12], and other physical sciences [13]. Some researchers have employed different modeling techniques to create classification models for the prediction of coronary heart diseases. Sellappan Palaniappan et al. proposed Naive Bayes, Neural Network, and Decision Trees models to create an Intelligent Heart Disease Prediction System (IHDPS) [14].

Other researchers used a tree classification method, Random Forest (RF), K-Nearest Neighbor (KNN), Support Vector Machine (SVM), Logistic Regression (LR) algorithms, to analyze the Framingham dataset [15]. When all these techniques were compared, the RF method performed better. On the well-known Cleveland heart dataset and Statlog datasets, some authors used machine learning (ML) algorithms to predict heart diseases based on some influential features. The results showed that the RF and SVM with grid search algorithms performed better on the Cleveland dataset, while the LR and NB classifiers performed better on the Statlog dataset [4].

## III. METHOD

### A. Dataset

In this study, a coronary heart disease dataset heart disease (Framingham dataset), which can be downloaded from Kaggle. This dataset comes from an ongoing cardiovascular study of Framingham, Massachusetts residents. The classification goal is to predict whether the patient will develop coronary heart disease in the next ten years (CHD). The dataset contains information about the patients. There are over 4,000 records and 16 attributes in total. Each characteristic is a potential risk factor. There are demographic, behavioral, and medical risk factors.

Kelvin Kwakye & Emmanuel Dadzie are PhD candidates at the Industrial and Systems Engineering Department of the North Carolina Agricultural and Technical State University, 1601 E Market St, Greensboro, NC 27401, USA (email:kkkwakye@aggies.ncat.edu, eadadzie@aggies.ncat.edu).

TABLE I

| Category | Description |
|---|---|
| **Demographic** | *Sex*: male or female (Nominal)<br>*Age*: Age of the patient;(Continuous - Although the recorded ages have been truncated to whole numbers, the concept of age is continuous)<br>*Education:* Educational background of the patient ranked from 1 to 4 (continuous) |
| **Behavioral** | *Current Smoker*: whether or not the patient is a current smoker (Nominal)<br>*Cigs Per Day*: the number of cigarettes that the person smoked on average in one day. (can be considered continuous as one can have any number of cigarettes, even half a cigarette.) |
| **Medical (history)** | *BP Meds*: whether or not the patient was on blood pressure medication (Nominal)<br>*Prevalent Stroke*: whether or not the patient had previously had a stroke (Nominal)<br>*Prevalent Hyp*: whether or not the patient was hypertensive (Nominal)<br>*Diabetes*: whether or not the patient had diabetes (Nominal) |
| **Medical (current)** | *Tot Chol*: total cholesterol level (Continuous)<br>Sys BP: systolic blood pressure (Continuous)<br>*Dia BP*: diastolic blood pressure (Continuous)<br>*BMI*: Body Mass Index (Continuous)<br>*Heart Rate*: heart rate (Continuous - In medical research, variables such as heart rate though in fact discrete, yet are considered continuous because of large number of possible values.)<br>*Glucose*: glucose level (Continuous) |
| **Predict variable (desired target)** | 10-year risk of coronary heart disease CHD (binary: "1", means "Yes", "0" means "No") |

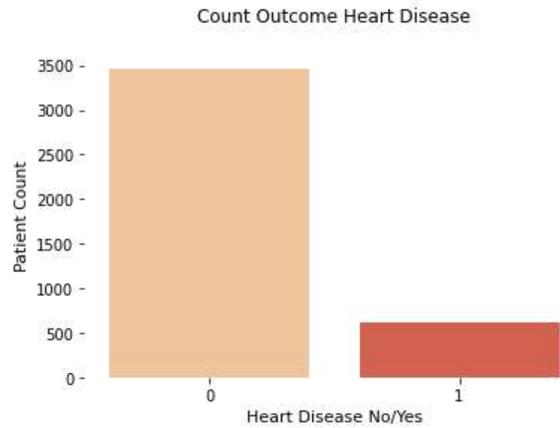

Fig 1. Distribution of the classes of the target variable

The distribution of the target variable which is shown in Fig 1. The target data which is a 10-year risk of coronary heart disease CHD has 3465 class of (cases) 0 and 617 class of 1. Therefore, this makes the dataset highly imbalance.

*B. Proposed Approach*

The current study's goal is to develop classification models for predicting coronary heart disease. The purpose of this research is to evaluate the performance of various classification models developed using various machine learning algorithms. The following paragraphs go over the specifics of the proposed approach.

First, the data is acquired from the repository and studied, which investigated the principles governing the instrument, the parameters, and attributes and the final measures from the setup. The data is examined for possible missing values representations and the range of each attribute. The next stage in our practice is preprocessing of the data. The categorical features are coded, missing values and outliers are identified and eliminated from the data. This stage is followed by feature engineering of the data to be used in model development. Fig 2 presents the structure of the proposed methodology.

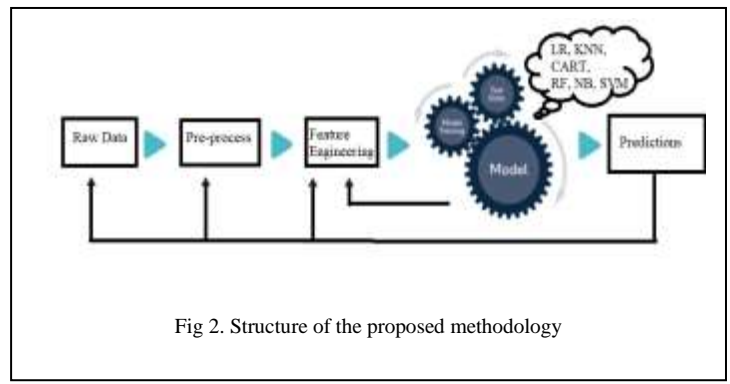

Fig 2. Structure of the proposed methodology

The feature engineering stage, which is the data transformation, entails both selecting the relevant features or attributes and completely modifying the data points. After this stage, various classification models are built using various machine learning algorithms.

Lastly, Cross-validation with 10-folds and the area under the curve (AUC) are two metrics used to assess each algorithm.

That is, the performance of the algorithms on the data is tested based on the cross-validation test and AUC separately.

*C. Preprocessing*

Data preprocessing began with the visualization of raw data using descriptive statistics tables, skewness, and other descriptions such as min, max, percentile values, and mean. It also includes the identification and removal of missing values, as well as the conversion of categorical values (the sex, Current smoker, and diabetes columns) into integers. The missing values in cigsPerDay, totChol, BMI, glucose, heartrate, were substituted with the mean values of each column. Also, the missing values of BPMeds which is categorical and education (ordinal with range 1-4) were removed from the dataset.

```
Sex                 0
age                 0
education         105
currentSmoker       0
cigsPerDay         29
BPMeds             53
prevalentStroke     0
prevalentHyp        0
diabetes            0
totChol            50
sysBP               0
diaBP               0
BMI                19
heartRate           1
glucose           388
TenYearCHD          0
dtype: int64
```

Fig 3. Number of Missing data

Fig 3 above shows the number of missing values found in each feature or attribute.

Further, outliers are removed from the data using the equations below.

$$Outlier > Q3 + 1.5 * (Q3 - Q1) \quad or \quad Outlier < Q1 - 1.5 * (Q3 - Q1) \quad (1)$$

$$Outlier - mean > |3 * s| \quad (2)$$

Where Q1, Q3, and s represent the first quantile, the third quantile, and the standard deviations of each attribute. A total of 697 outliers were identified. Equation 2 found 3 times as many outliers as Eq 1. Hence, Eq 2 was used to clean the data. From the box plot in Fig 4, outliers can be found in the following columns: cigsPerDay, totalChol, sysBP, diaBP, BMI, heartRate, and glucose. Although there are extremes in 'totChol' and 'sysBP,' but the majority of the outliers are close to the upper whisker, which is significant.

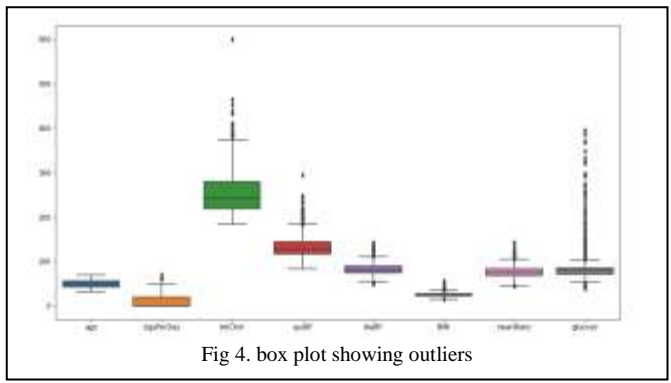

Fig 4. box plot showing outliers

```
Sex                 0
age                 0
education           0
currentSmoker       0
cigsPerDay         11
BPMeds              0
prevalentStroke     0
prevalentHyp        0
diabetes            0
totChol            54
sysBP             127
diaBP              84
BMI                96
heartRate          76
glucose           249
TenYearCHD          0
```

Fig 5. Number of outliers identified

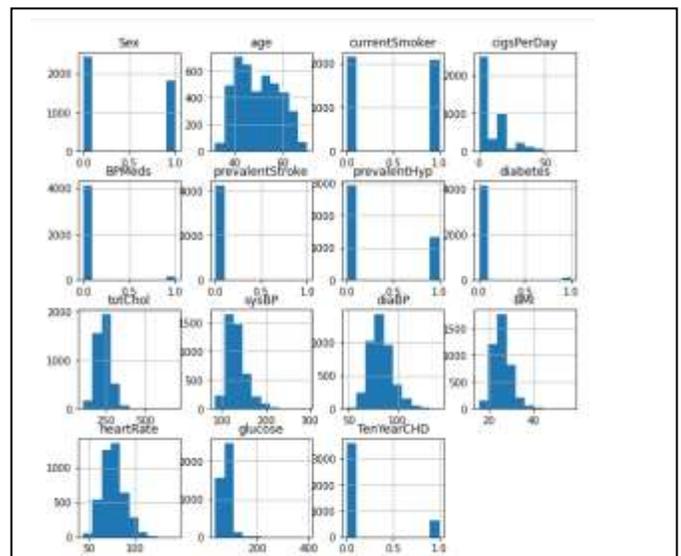

Fig 6. Histogram distribution of all the features

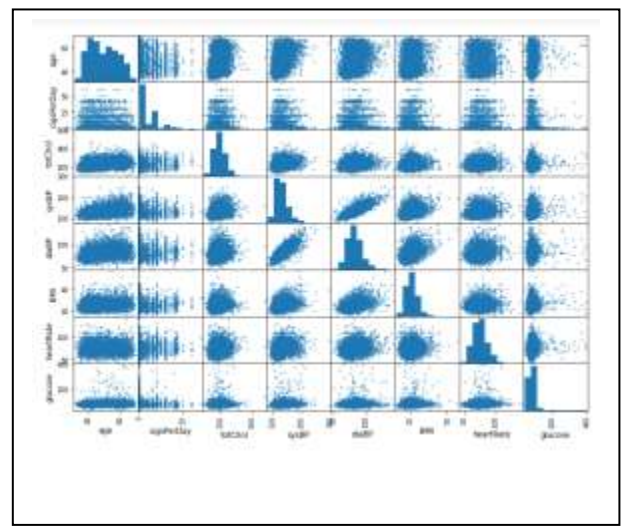

Fig 7. Scatter matrix of the numerical features

Fig 5 reports the number of outliers found. Fig 6 shows the histogram distribution of all the features and Fig 7 is a scatter matrix of the numerical features.

*Correlation Matrix*

The correlation matrix illustrates how the features are related to one another or to the target variable. The correlation heatmap in Fig 8 (a) shows that sysBP and diaBP are highly correlated, as are currentSmoker and cigsPerDay. Fig 8 (b) is the correlation matrix for the numerical features of the dataset.

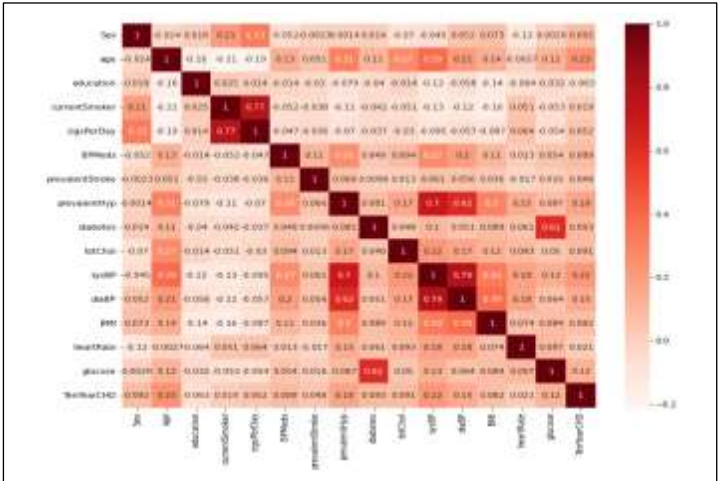

Fig 8 (a) correlation heatmap of all features

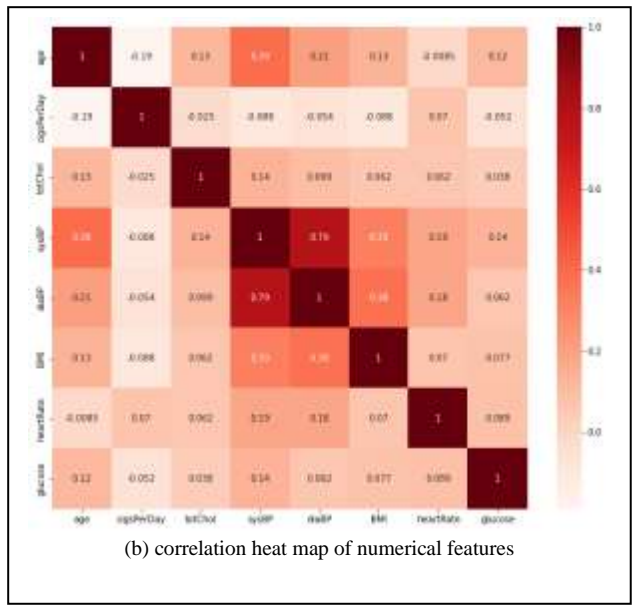

(b) correlation heat map of numerical features

### D. Feature Engineering

This section of the study entails both the selection of relevant features from the group of attributes and the extraction of hidden features within the attributes that account for more than 95% of the variance in the data. A filter-based feature selection method was used on the dataset. It uses statistical measures to score the correlation or dependence between input variables, which can then be filtered to select the most relevant features. Specifically, it considered the Mutual Information Feature Selection feature extraction or selection technique which employs SelectKBest feature selection strategy. It computes between two variables and measures the reduction in uncertainty for one variable given a known value of the other variable [16]. Fig 9 and Fig 10 shows the scores and plot of the relevant features respectively.

```
Feature 0:  0.025866
Feature 1:  0.000000
Feature 2:  0.008760
Feature 3:  0.012026
Feature 4:  0.015536
Feature 5:  0.004695
Feature 6:  0.008744
Feature 7:  0.000000
Feature 8:  0.014122
Feature 9:  0.004497
Feature 10: 0.002761
Feature 11: 0.003122
Feature 12: 0.004135
Feature 13: 0.012066
Feature 14: 0.007720
```

Fig 9. feature scores

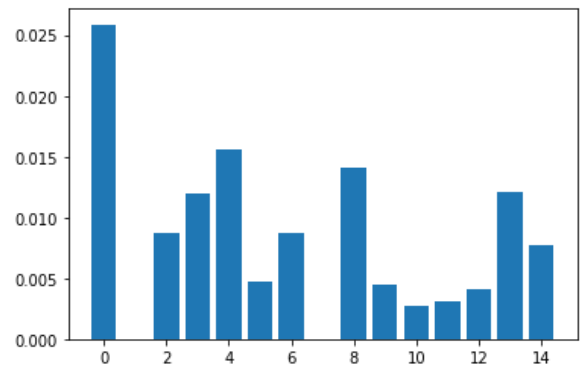

Fig 10. plot showing the feature scores

### E. Classification Model Development

The final stage entails the development of a model using various machine learning algorithms. This project's algorithms include k-nearest neighbor (KNN), support vector machines (SVM), decision tree (CART), logistic regression (LR), nave bayes (NB), and Random Forest (RF). Following feature selection, the dataset was divided into training and test sets (80:20), which were trained using the aforementioned algorithms. The performance of the models was then assessed using two different metrics: cross-validation accuracy and the area under the curve (AUC) of the receiver operating characteristics (ROC).

## IV. RESULTS

The current application starts with data preparation, which replaces and removes unwanted data points that could introduce biases into the classification model. The data preparation also aims to improve the models' performance.

### A. Data Preprocessing

*1) Replacing missing values and outliers*

The cleaning of the data starts with the replacing of missing values with their mean values. The mean values of each column were used to replace missing values in cigsPerDay, totalChol, BMI, glucose, and heartrate. In addition, the missing values of BPMeds that are categorical and education (ordinal with range or rank 1-4) were completely removed to ensure that there was no missing data in the dataset. Outliers that existed were removed as shown in Fig 11.

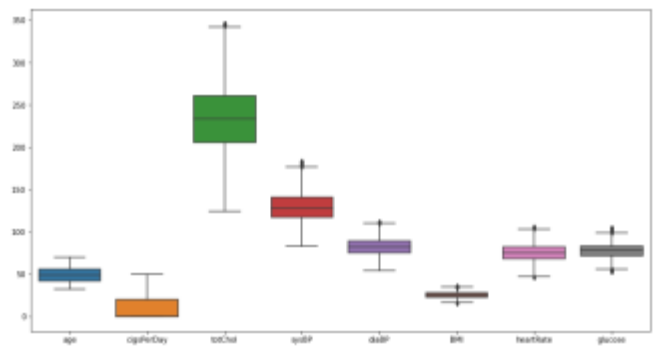

Fig 11. Box plot showing results after removing the outliers

The visualization of the data after cleaning (the removal of outliers) is critical for determining any potential improvements in the distributions and patterns. Fig 12 and Fig 13 display scatter matrix and histogram plots of the cleaned dataset showing distinct patterns and distributions across the attributes, which indicates the accurate description of the data. When those values and the outlier were removed, the attributes returned to normal, with skewness values approaching zero. Furthermore, the scatter matrix visualization demonstrates that the dataset with the unwanted values produced plots with unrecognizable patterns. However, after cleaning the data, the plots produced distinct patterns for each pair of attributes, as well as near-normal histograms for each attribute.

The skewness values for the attributes now indicate a normal or near-normal distribution (close to zero). Furthermore, the clean dataset's correlation values are more realistic.

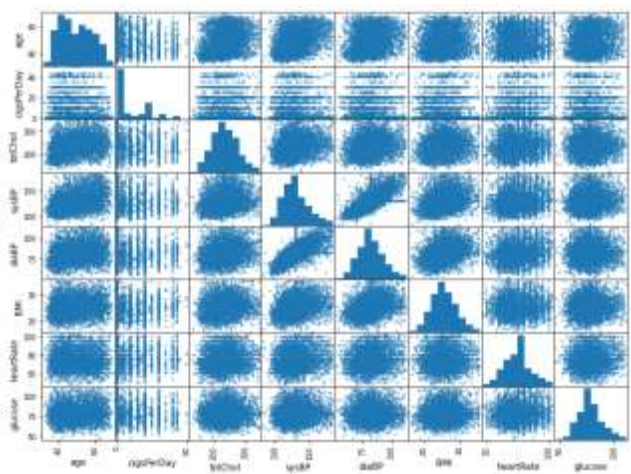

Fig 12. The scatter matrix for the dataset after the removal of outliers. The diagonal of the matrix shows the histogram plot for each attribute

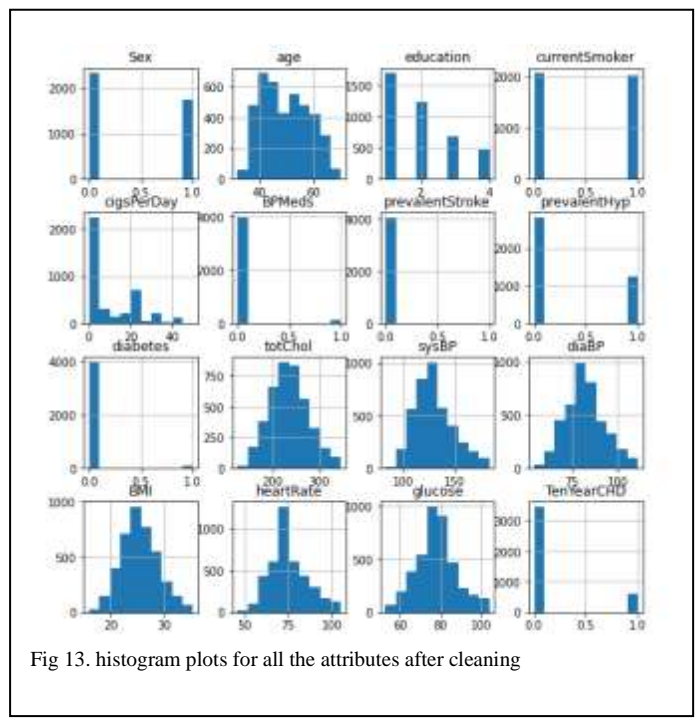

Fig 13. histogram plots for all the attributes after cleaning

B. Machine learning classification models

1) Outcome of the cross-validation (CV) test

The results for the original dataset revealed extremely low mean accuracy, particularly in the SVM model. As a result, the data was transformed using the Synthetic Minority Oversampling Technique (SMOTE), which increased the number of cases in the dataset in a balanced manner. That is, it created new instances from existing minority cases. This transformation produced higher mean accuracy. Table II and Table III presents the summary results for the CV-evaluated model output.

TABLE II
SUMMARY RESULTS OF CROSS-VALIDATION TEST BASED ON ROC-AUC FOR UNBALANCE-DATA ALGORITHM MODELS

| Parameter | LR | KNN | CART | NB | SVM | RF |
|---|---|---|---|---|---|---|
| ORIGINAL | | | | | | |
| Mean | 0.728592 | 0.599152 | 0.543827 | 0.707762 | 0.521957 | 0.686704 |
| Std | 0.030915 | 0.049377 | 0.017969 | 0.034112 | 0.078160 | 0.034263 |

TABLE III
SUMMARY RESULTS OF CROSS-VALIDATION TEST BASED ON ROC-AUC FOR BALANCE-DATA ALGORITHM MODELS

| Parameter | LR | KNN | CART | NB | SVM | RF |
|---|---|---|---|---|---|---|
| SMOTE | | | | | | |
| Mean | 0.729461 | 0.886542 | 0.784070 | 0.721225 | 0.655441 | 0.946337 |
| Std | 0.020252 | 0.011446 | 0.010890 | 0.028593 | 0.028593 | 0.006369 |

### 2) Outcome of Prediction Test

The results show that the values are relatively consistent, with similar variability across the different transformations. Table IV summarizes the ROC-AUC-evaluated model output for the original (unbalanced) and smote datasets (balanced)

TABLE IV
SUMMARY RESULTS OF PREDICTION TEST BASED ON ROC-AUC FOR BOTH BALANCED- AND UNBALANCE-DATA ALGORITHM MODELS

| Data | LR | KNN | CART | NB | SVM | RF |
|---|---|---|---|---|---|---|
| Original | 0.5127 | 0.5131 | 0.5233 | 0.5474 | 0.5 | 0.4995 |
| Smote | 0.6747 | 0.7794 | 0.7859 | 0.6653 | 0.5 | 0.8769 |

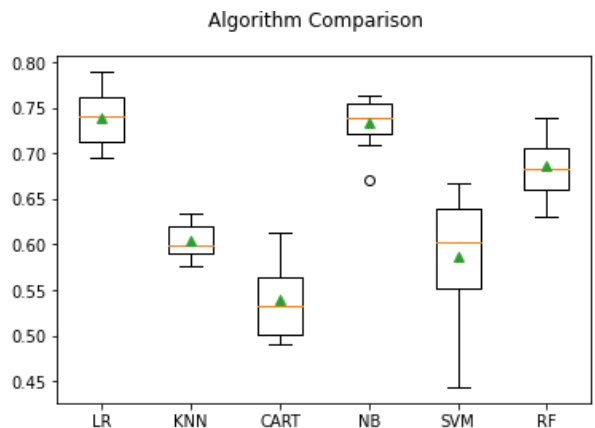

Fig 14. Box plot of the various algorithms with original Data

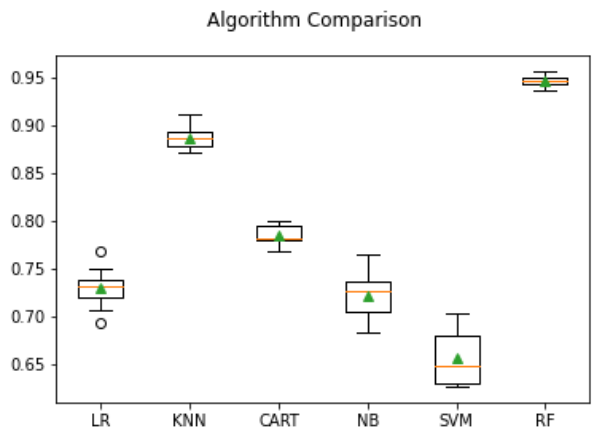

Fig 15. Box plot of the various algorithms with Smote Data

## V. DISCUSSION AND CONCLUSIONS

The present study developed multiple classification models for the prediction of coronary heart disease using different machine learning algorithms. After building the models with the original data set, the performance of the various algorithms were relatively low due to the target feature being highly unbalanced. A Smote was introduced to transform the initial data by balancing the classes. Two performance metrics, cross-validation accuracy, and Hold-out prediction (HOP) testing using ROC_AUC were used to evaluate the performance of the models. The CV helps to determine the consistency of the model and that also indicate the variance of the model. The HOP reveals the prediction power of the model against sample not previously seem by the model. Hence, the models estimated performance to real world samples. By comparing the CV and HOP, a model can be determined to either overfit the training data sample or not.

The results indicated that at default settings, For the unbalance data models, the LR and the NB outperformed the other models with **0.728592** and **0.707762** accuracy respectively. Their variance were relatively lower than the remaining algorithms. Surprisingly, the linear model, that is LR, was the algorithms with the highest CV performance which is unexpected for such a classification problem. Usually, non-linear models do better in this problem scenarios. The CART models produced the worst accuracy among the models.

The balance dataset, on the other hand, fixed the class bias within the dataset and allow a more realistic models to be develop. In this case, most of the non-linear models outperformed the linear model, that is, the LR. The random forest (RF), which is an ensemble model outperformed all the remaining algorithms with an accuracy value of **0.946337**. The next best performing algorithm model was KNN with a CV accuracy of **0.886542**. The SVM was the weakest models in this model scenario. In comparison, the balanced-data models had relatively lower variance in their results than the unbalance-data models.

For the Prediction test, in reference to the unbalance-data models, the NB model produced the highest prediction accuracy with a value of **0.5474**. This was followed by CART and KNN. When the CV AUC values are compared to the prediction AUC values for the unbalance-data models, the results indicate there is a general overfitting among all the models. This is because the difference between the CV and the predictions values are large.

For the balanced-data models, the RF still outperformed all the other models, followed by CART and KNN. The juxtaposition of the CV and prediction AUC values indicate well fitted models among most of algorithms

Hyper-parameter tuning of the highest-performing model was done and it did not produce much difference in performance compared to the default setting. Therefore, the default logistic regression algorithm is the best machine learning model for this problem of coronary heart disease prediction.

In conclusion, this study suggests that LR on a well-processed and standardized dataset can predict coronary heart disease with greater accuracy than the other algorithms

## VI. FUTURE WORKS

In the future, missing values and outliers would be investigated using imputation techniques and other methods that avoid the use of a single value such as the mean for data cleaning and pre-processing. These methods are expected to prevent the narrowing of the probability distribution that will prevent the data points close to the outlier limits from being pushed beyond the limit. Hence, alleviating the further emergence of new outliers after the data is preprocessed.

Furthermore, all of the algorithms or, at the very least, the best performing models can be combined into an ensemble to create unified models that may improve prediction performance.